\documentclass[runningheads]{llncs}
\usepackage{graphicx}
\usepackage{amsmath,amssymb,amsfonts}
\usepackage{float}
\usepackage{subcaption}
\usepackage{multirow}

\DeclareMathOperator{\Tr}{Tr}
\usepackage{algorithm}
\usepackage{algpseudocode}
\algrenewcommand\algorithmicrequire{\textbf{Input:}}
\algrenewcommand\algorithmicensure{\textbf{Output:}}

\DeclareMathOperator*{\argmin}{arg\,min}

\usepackage[misc]{ifsym}

\begin{document}
\title{Manufacturing Dispatching using Reinforcement\\and Transfer Learning}

\titlerunning{Manufacturing Dispatching using Reinforcement and Transfer Learning}
%\author{}
\author{Shuai Zheng\textsuperscript{(\Letter)} \and Chetan Gupta \and Susumu Serita}
%First Author\inst{1}\orcidID{0000-1111-2222-3333} \and
%Second Author\inst{2,3}\orcidID{1111-2222-3333-4444} \and
%Third Author\inst{3}\orcidID{2222--3333-4444-5555}}
%
\authorrunning{S. Zheng et al.}
\institute{Industrial AI Lab, Hitachi America Ltd\\Santa Clara, CA, USA\\
\email{\{Shuai.Zheng,Chetan.Gupta,Susumu.Serita\}@hal.hitachi.com}}

\tocauthor{Shuai Zheng, Chetan Gupta, Susumu Serita}
\toctitle{Manufacturing Dispatching using Reinforcement and Transfer Learning}

%Princeton University, Princeton NJ 08544, USA \and
%Springer Heidelberg, Tiergartenstr. 17, 69121 Heidelberg, Germany
\maketitle              % typeset the header of the contribution
\begin{abstract}
Efficient dispatching rule in manufacturing industry is key to ensure product on-time delivery and minimum past-due and inventory cost. Manufacturing, especially in the developed world, is moving towards on-demand manufacturing meaning a high mix, low volume product mix. This requires efficient dispatching that can work in dynamic and stochastic environments, meaning it allows for quick response to new orders received and can work over a disparate set of shop floor settings.  In this paper we address this problem of dispatching in manufacturing. Using reinforcement learning (RL), we propose a new design to formulate the shop floor state as a 2-D matrix, incorporate job slack time into state representation, and design lateness and tardiness rewards function for dispatching purpose. However, maintaining a separate RL model for each production line on a manufacturing shop floor is costly and often infeasible. To address this, we enhance our deep RL model with an approach for dispatching policy transfer. This increases policy generalization and saves time and cost for model training and data collection. Experiments show that: (1) our approach performs the best in terms of total discounted reward and average lateness, tardiness, (2) the proposed policy transfer approach reduces training time and increases policy generalization.
\end{abstract}

\section{Introduction}

In a manufacturing process, a production order moves through a sequence of job processing steps to arrive at a final product. The problem of dispatching is the assigning the next job to be processed for a given machine. Inefficient scheduling and dispatching can cause past-due cost (past-due cost is the cost when a job cannot be delivered on time) as well as inventory cost (inventory cost is the storage cost when the job is finished before due time) to go up. It is obvious that all manufacturing managers want on-time delivery and minimum past-due and inventory cost. However, achieving these goals requires efficient production scheduling that minimizes these costs. Furthermore, manufacturing is moving towards a high mix low volume product mix, which makes this even more challenging, due to an ever-evolving product mix, causing larger variations in job types and job arrival rates.

Production scheduling problems can be categorized by shop configurations and scheduling objectives. Depending on the number of processing machines, there can be single-machine environment \cite{park2013genetic} and parallel-machine environment \cite{jakobovic2007genetic}.
Depending on the number of operations (or stages, steps) of each job, there are single-operation and multi-operation environment \cite{garey1976complexity}. Depending on the objective of scheduling, there are completion time based scheduling \cite{vazquez2011automatic} (trying to increase machine efficiency) and due date based scheduling \cite{mascia2013grammars} (trying to be close to promised delivery dates). Multi-operation parallel-machine problems can be solved through multi-agent algorithms or can be decomposed into solving several single-operation problems \cite{branke2016automated}. In this work, we focus on dynamic dispatching for due date based objective, which has broader generalization and can be used in different shop floor settings. Dynamic dispatching is also critical for predictive maintenance tasks \cite{mobley2002introduction,zheng2017long}, which schedule machine maintenance using data-driven and model-based approaches. Furthermore, dispatching only schedules the imminent job with the highest priority and is particularly suitable for dynamic/stochastic environments due to its low computational time in deployment/testing stage.

Traditionally to address the problem of dispatching, a lot of hyper-heuristics have been proposed and shown to be effective and reusable in different shop conditions \cite{branke2016automated}. Even though some exact solutions for deterministic scheduling problems are available, manufacturing shop floor depends on heuristics. This is because exact solutions are computationally expensive (and hence infeasible) in deployment stage and cannot solve problems for dynamic and stochastic environments. Since heuristics are very problem-specific and usually achieved by trial and error, hyper-heuristics \cite{branke2016automated} which automate the design of heuristics attract a lot of interest and have been shown effective in manufacturing as well as other industries such as: bin packing \cite{ozcan2011policy}, vehicle routing \cite{vonolfen2013structural}, project scheduling \cite{frankola2008evolutionary}. Many hyper-heuristics are based on machine learning approaches, such as neural networks, logistic regression, decision trees, Support Vector Machines, genetic algorithms, genetic programming, reinforcement learning.  

Recently, existing works have used deep reinforcement learning for scheduling problems \cite{mao2016resource,chen2010rule}. For dispatching purpose, we propose a new design to formulate the shop floor state as a 2-D matrix, incorporate job slack time into state representation. We also design lateness and tardiness rewards function for reinforcement learning. In a manufacturing shop, there are many similar production lines, where designing and maintaining a separate reinforcement learning model for each line is not feasible. To address this, we propose a transfer approach for dispatching policy using manifold alignment. 
Unlike discriminant subspace learning \cite{zheng2014kernel,zheng2016harmonic,zheng2018harmonic} and sparse enforced learning \cite{zheng2019sparse,zheng2018minimal,zheng2018regularized} where the purpose is to separate classes, manifold alignment \cite{wang2009manifold} learns a subspace by matching the local geometry and preserving the neighborhood relationship within each space. 
In summary, the contributions of this work are:
\begin{enumerate}
\item The reinforcement learning module uses deep learning and policy gradient to minimize job due related cost. Compared to existing work, the novelty is that we formulate the shop floor state as a 2-D matrix, incorporate job slack time into state representation, and design lateness and tardiness rewards function for dispatching purpose. 
\item The transfer learning module transfers dispatching policy between shop floors using manifold alignment. Compared to existing work, the novelty is that our approach formulates shop floor states transfer using manifold alignment, and we design a method to recover actions from sequence of states.
\end{enumerate}

To test our approach, we build a simulator to simulate dynamic factory settings, where we can change factory configurations and job characteristics, such as job queue length, machine processing capacity, job arrival distribution. A simulator is needed since it is very difficult to run live experiments on an actual production line. Another option is to somehow use actual data in a shop floor, however this is a challenge: the data collected on the shop floor is mainly state and scheduling decision record data which is static in nature and cannot provide feedback on the scheduling decisions. These data can be used for hyper-heuristics rules extraction, but not for \emph{dynamic} environment \cite{branke2016automated}. Furthermore, corporations are reluctant to share real data, because these data include trade secrets and sensitive financial information. In fact, most published works for \emph{dynamic} production scheduling are based on simulators \cite{branke2016automated}. Using simulator to pre-train model and then deploying it into real factories can further reduce data collection and training effort.

\section{Related work}

\subsection{Production scheduling}

Hyper-heuristics are promising to handle dynamic and stochastic scheduling problems and have recently emerged as a powerful approach to automate the design of heuristics for production scheduling \cite{branke2016automated}. The basic idea of hyper-heuristics is to learn scheduling rules from a set of very good training scheduling instances. These training instances are considered optimal or perfect. Hyper-heuristics can then replicate these scheduling instances as closely as possible. Depending on the learning method, hyper-heuristics can be classified into supervised learning and unsupervised learning. Examples of supervised learning hyper-heuristics include neural networks \cite{weckman2008neural}, logistic regression \cite{ingimundardottir2011supervised}, decision trees \cite{li2005discovering}, Support Vector Machines \cite{shiue2009data}, etc. Genetic algorithms (GA) and Genetic Programming (GP) are evolutionary computation methods and have been used for dynamic job shop scheduling problems. Reinforcement learning is an efficient algorithm to learn optimal behaviors through reward feedback information from dynamic environments \cite{chen2010rule}. $TD(\lambda)$ based reinforcement learning was used for manufacturing job shop scheduling to improve resource utilization \cite{zhang1995reinforcement}. Resource scheduling for computer clusters is also very related, such as Tetris \cite{grandl2015multi} and resource scheduling in HPC \cite{zheng2011analysis}. RL-Mao \cite{mao2016resource} uses policy gradient reinforcement learning to reduce computer job slowdown. The difference of our work and RL-Mao lies in that our work integrates manufacturing job slack time into state representation and the objective functions using lateness and tardiness are specifically designed for manufacturing dispatching. 

\subsection{Reinforcement learning background}
In reinforcement learning, an agent interacts with an environment $\mathcal{E}$ over many discrete time steps \cite{sutton1998reinforcement}. The state space of $\mathcal{E}$ is defined within $\mathcal{S}$. At each time step $t$, the agent receives a state $s_t \in \mathcal{S}$ and performs an action $a_t \in \mathcal{A}$ following a policy $\pi$, where $\mathcal{A}$ defines the action space of this agent. The agent receives a reward $r_t$ for this action and a new state $s_{t+1}$ is then presented to the agent. The policy $\pi$ is a mapping function from states $s_t$ to $a_t$, denoted by $\pi(a_t | s_t)$, which gives the probability of taking action $a_t$. This process continues until the agent reaches a termination state or time $t$ exceeds a maximum threshold. The cumulative discounted reward starting from time $t$ is defined as: 
\begin{align}
R_t = \sum_{k=0}^\infty \gamma^{k} r_{t+k}, \label{eq:r}
\end{align}
where $\gamma \in (0,1]$ is a discounted factor. The goal of a reinforcement agent is to obtain a policy which maximizes the expected total discounted reward starting from time $t=0$:
\begin{align}
J(\pi) = \mathbb{E}[ R_0|\pi]. \label{eq:j}
\end{align}

In policy-based model-free reinforcement learning, policy is directly parameterized as a function from states to actions, $\pi(a|s;\theta)$, where parameter $\theta$ is updated using gradient ascent on $\mathbb{E}[R_t|\pi]$. One example of this category is the REINFORCE algorithm \cite{williams1992simple,sutton2000policy}. Using the policy gradient theorem \cite{williams1992simple}, the gradient with respect to $\theta$ can be given as $\nabla_\theta \log \pi(a_t | s_t;\theta)R_t$, which is an unbiased estimate of $\nabla_\theta \mathbb{E}[R_t|\pi]$. In order to reduce the variance of this estimate, we can subtract a baseline from the return, where baseline $b_t(s_t)$ is a learned function of state. The resulting gradient is thus given as 
\begin{align}
\nabla_\theta \log \pi(a_t | s_t;\theta) (R_t-b_t).
\label{eq:grad}
\end{align}
The term $R_t - b_t$ is used to scale the policy gradient and can be seen as the advantage of action $a_t$ in state $s_t$.

\begin{figure}[t]
\centering
\includegraphics[width=0.5\columnwidth]{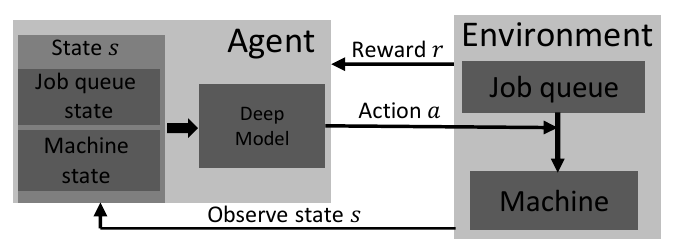}
\caption{Overall design of Deep Manufacturing Dispatching (DMD).}
\label{fig:reinforcement}
\end{figure}

\section{Dispatching with reinforcement learning}

\subsection{Problem description}

\textbf{Objective.} Dispatching performance can be evaluated in terms of lateness $L$ and tardiness $TA$. The objective of this problem is to minimize the average lateness and tardiness of all jobs. For one single job, let the completion time be $c$. Lateness is the absolute difference between job due time and job completion time: 
\begin{align}
L = | c - d |, \label{eq:lateness}
\end{align}
where $L\in[0, +\infty)$. Thus lateness considers both inventory and past-due cost. Tardiness only considers the lateness when the job is late. Tardiness $TA$ is defined as: 
\begin{align}
TA = max(c - d, 0), \label{eq:tardiness}
\end{align}
where $TA\in[0, +\infty)$. Tardiness focuses on past-due cost, but not inventory cost.

\textbf{Constraints.}  
Since this is a single-operation environment, there is no precedence constraint between jobs/operations. Disjunctive constraint is enforced in this problem, so that no two jobs can be processed at the same time. If the job is being processed, it cannot be paused. All jobs are equally important.   

\subsection{Design}

Figure \ref{fig:reinforcement} shows the overall design of proposed Deep Manufacturing Dispatching (DMD). The environment includes job queue and processing machine. At each time step, reinforcement learning agent observes a state $s$, which includes job queue state and machine state (schedule of next $T$ time steps), then outputs a probability vector with respect to each action using a deep learning model as function approximator. The agent will then perform action $a$ with the highest probability and receive a reward $r$ from the environment. 

There are $n$ job slots and $m$ backlog slots. Each job has a processing time (job length) $p$ and a due time $d$. At each time step, the probability of arriving a new job is $\lambda \in (0, 1)$. When a job arrives, it will be placed on one of the $n$ job slots randomly. If job slots are full, the job will be placed on backlog slots. For jobs placed on backlog slots, only job count can be seen and those jobs cannot be selected by dispatcher. As the backlog design for computer jobs in \cite{mao2016resource}, it is reasonable that the reinforcement agent only considers jobs in job slots, not those on backlog slots, because jobs in job slots arrive earlier and get higher priority. Let $t_{curr}$ indicate current time. Slack time $slack$ of a job is defined as:
\begin{align}
slack = d - t_{curr} - p. \label{eq:slack}
\end{align}
If $slack>0$, it means that if this job is started now, it will be completed before its due time; if $slack<0$, it means that it will be completed after its due time. We now explain the design details of DMD elements: state space, action space, reward, training.

\begin{figure}[t]
\centering
\includegraphics[width=0.5\columnwidth]{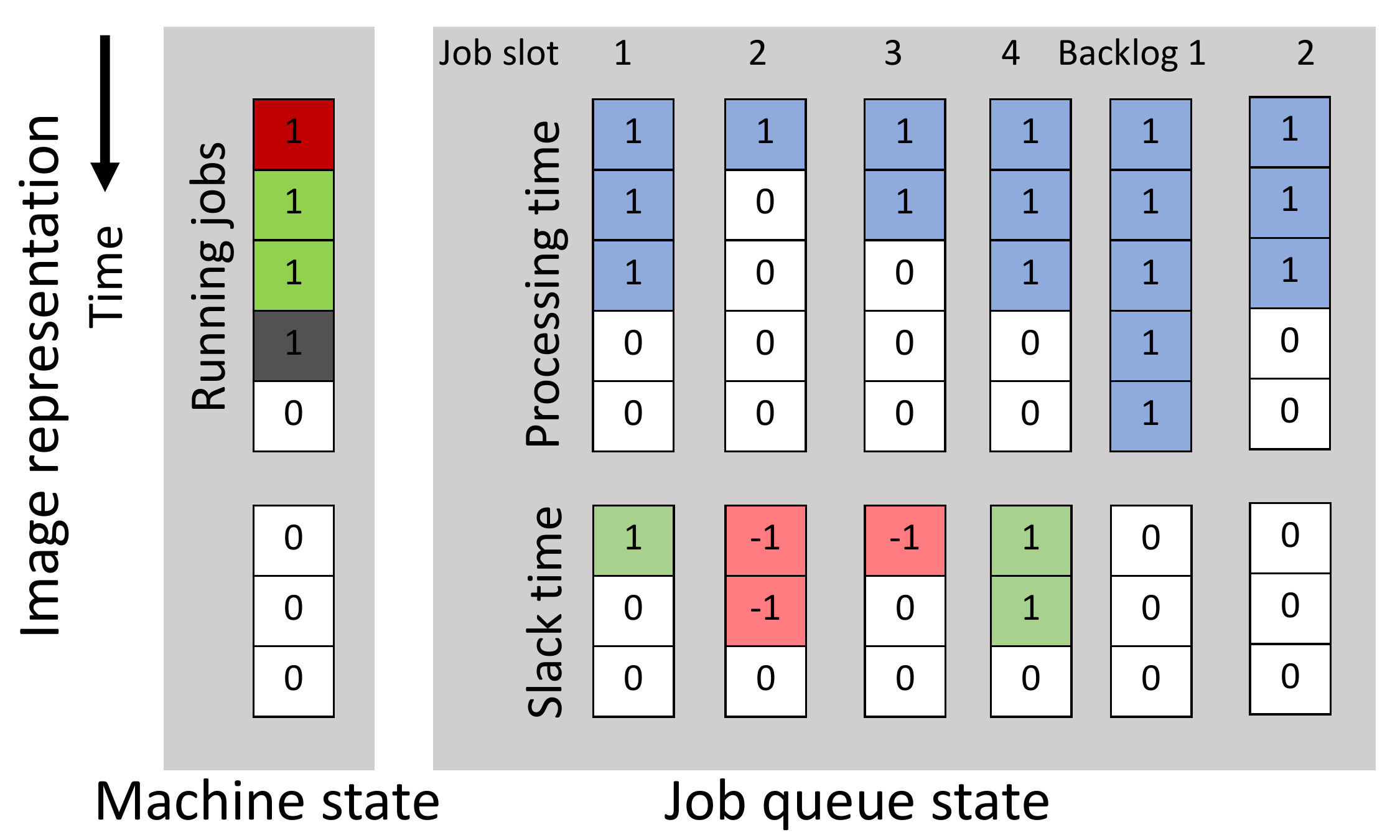}
\caption{State representation: this state looks $T=5$ time steps ahead of current time and slack time array length is $Z=3$; there are $n=4$ job slots and $m=10$ backlog slots. The machine state (schedule of next $T$ time steps) tells that the next $4$ time steps have been scheduled. For the job in job slot 1, $p=3$, $slack = 1$; for the job in job slot 3, $p=2$, $slack = -1$. There are $8$ jobs waiting in backlog. Different colors in machine state represent different jobs.}
\label{fig:image}
\end{figure}

\textbf{State space}. State space includes both states of the machine and job queue. At any time step, we consider the states $T$ time steps ahead and we use a $Z$ length array to represent the slack time. We use a 2-D matrix to represent machine state (schedule of next $T$ time steps) and job queue state. One example is shown in Figure \ref{fig:image}. Value $1$ in machine state means that the machine at that time step has been allocated, and value $0$ means that the machine will be idle. As time proceeds one time step, the $T$-length array shifts up by one unit and a new $0$ value is appended to the bottom. Job queue state consists of $n$ job slots and $m$ backlog. Different colors in machine state indicate different jobs.

Each job slot is represented using two arrays: processing time and slack time. The processing time array is represented using a $T$ length array, with the number of $1$s indicating job length $p$. The slack time is represented using a $Z$ length array, where $1$ means positive $slack$ and $-1$ means negative $slack$. The sum of slack time array represents job $slack$. 

Backlog state is represented by several $T$ length array, with a total of $m$ slots, where each slot represents a job. In Figure \ref{fig:image}, backlog is represented using $2$ 5-length array and there are 8 jobs in backlog. The $0$s under machine state and backlog do not represent slack time and are padded to make the state as a complete 2-D matrix. There are 2 benefits using this 2-D representation: 1). 2-D representation can capture the relationship between job characteristics and machine status, and we can use deep learning method to discover hidden patterns from 2-D representations; 2). we can represent schedule in $T$ time steps ahead of time.

\textbf{Action space}. At each time step, the dispatcher performs multiple actions to select a subset of jobs from $n$ job slots until an invalid or void action is selected. The action $a$ is a subset of $\{\emptyset, 1, ...,n\}$, where $\emptyset$ is a void action and no job slot is selected. Void action allows dispatcher to save more resources for larger jobs.

We let the dispatcher make a decision at each time step, instead of only when the machine is idle. This is because jobs are scheduled $T$ time steps ahead. The decision of the dispatcher is to decide which job to allocate within the next $T$ time step, not for which job to run next immediately, so we do not consider the machine is idle or not. Instead, we consider the schedule $T$ time steps ahead.

\textbf{Reward}. We design the reward at each time step with respect to lateness and tardiness as:
\begin{align}
r_L &= -\sum_{j \in \mathbb{P} }\frac{L_j}{p_j},\label{eq:rl}
\\
r_{TA} &= -\sum_{j \in \mathbb{P} }\frac{TA_j}{p_j},\label{eq:rta}
\end{align}
where $\mathbb{P}$ is set of jobs that are currently being processed by the machine. Summation of $r_L$ and $r_{TA}$ over all time steps for the running jobs equals to the total lateness and tardiness.

\textbf{Training}. We run the simulator multiple times to get a batch of trajectories. At each time $t$, we record the state $s_t$, action $a_t$ and reward $r_t$. The discounted reward $R_t$ at time $t$ can then be computed using Eq.(\ref{eq:r}). $b_t$ is a baseline function, which can be the average reward at time $t$ of multiple trajectories.

\subsection{Scalability and generalization}
The complexity comes from training and deployment/testing stage. In training stage, the computational time mainly depends on the training of reinforcement learning agent. From Figure \ref{fig:slack}, \ref{fig:transfer}, we found that the objective function converges fast. Pre-training with simulated data, transfer learning and other accelerated computing \cite{zheng2016accelerating} can further reduce training time. In deployment/testing stage, the computational time is very fast and the same as testing time of a deep neural network. This ensures the quick response to different shop floor conditions.

\section{Dispatching policy transfer}
In reinforcement learning area, some approaches using manifold learning for cross-domain transfer learning \cite{ammar2015unsupervised,joshi2018cross,pan2010survey} have been applied in games and apprentice learning, where control of games, such as Cart Pole and Simple Mass, shares similar control mechanism to balance an object. However, for dispatching problems, due to the complexity of manufacturing settings, it is not straightforward to transfer dispatching rules among factories or product lines. In our proposed Deep Manufacturing Dispatching (DMD) framework, the following features of the training data will affect the data distribution: 1. factory setting parameters, which are used to describe job queue states and machine states, including length of processing time array $T$, length of slack time array $Z$, number of job slots $n$, number of backlog slots $m$, etc.; 2. job characteristics parameters, such as job length distribution, job arrival speed, job due time distribution. To apply a trained policy in a new factory setting or when job characteristics changes, knowledge transfer would greatly improve the performance of learning by avoiding expensive data collection process and reducing training time. 

Given source environment $\mathcal{E}_x$ with state space $\mathcal{S}^x$ and action space $\mathcal{A}^x$, target environment $\mathcal{E}_y$ with state space $\mathcal{S}^y$ and action space $\mathcal{A}^y$, source optimal policy $\pi^{x\ast}(a|s)$ is already learned. Transfer learning can be used to learn target optimal policy $\pi^{y\ast}(a|s)$. There are two types of policy transfer: (1) same-environment transfer, where factory setting parameters are not changed, but job characteristics parameters are changed; (2) cross-environment transfer, where factory setting parameters are changed. For cross-environment transfer, the job queue and machine states are changed, and input dimension of source policy is different from the input dimension of target policy, so source policy cannot be applied directly in new environment. To successfully and effectively transfer $\pi^{x\ast}$ to $\pi^{y\ast}$, we propose the following transfer strategy as shown in Algorithms \ref{alg:alg1}. 
 
In step 1, we want to find a state projection $\chi$, so that for any source state $s^x \in \mathcal{S}^x$, a corresponding target state is given as:
\begin{align}
s^y=\chi s^x. \label{eq:chi}
\end{align}
We will introduce how to find this projection $\chi$ in Algorithm \ref{alg:alg2}. Using computed target environment trajectories from step 3, step 4 recovers policy $\pi^{y}$ using Algorithm \ref{alg:alg3}.

\begin{algorithm}[t]
\caption{Policy transfer learning.}
\label{alg:alg1}
\begin{algorithmic}[1]
\Require Source environment $\mathcal{E}_x$, target environment $\mathcal{E}_y$, source optimal policy $\pi^{x\ast}$.
\Ensure Target optimal policy $\pi^{y\ast}$.
\State Find state projection $\chi$: $\mathcal{S}^x \to \mathcal{S}^y$ using Algorithm \ref{alg:alg2}. 
\State Following optimal policy $\pi^{x\ast}$, generate source state trajectories, $\{s^x_{00}, s^x_{01}, ...\}$, $\{s^x_{10}, s^x_{11}, ...\}$, ... 
\State Compute target state trajectories in $\mathcal{S}^y$, $\{ s^y_{00}, s^y_{01}, ... \}$, $\{ s^y_{10}, s^y_{11}, ... \}$,..., using Eq.(\ref{eq:chi}). 
\State Recover policy $\pi^{y}$ using Algorithm \ref{alg:alg3}.
\State Fine-tune policy $\pi^{y}$ and get optimal policy $\pi^{y\ast}$.  
\end{algorithmic}
\end{algorithm}

\textbf{Find projection $\chi$}. 
Manifold alignment \cite{wang2009manifold} learns a subspace by matching the local geometry within source space and target space. Given some random source states $s^x_0, s^x_1,...$, random target states $s^y_0, s^y_1,...$, manifold alignment looks for two projections $\alpha$ and $\beta$ to minimize cost function:
\begin{align}
C(\alpha, \beta) = &\mu \sum_{i,j} (\alpha^T s_i^x - \beta^T s_j^y)^2 W^{i,j} +0.5 \sum_{i,j}(\alpha^T s_i^x - \alpha^T s_j^x)^2 W_x^{i,j} \nonumber  \\
& + 0.5 \sum_{i,j}(\beta^T s_i^y - \beta^T s_j^y)^2 W_y^{i,j},
\label{eq:cost}
\end{align}
where $\alpha$ and $\beta$ project source states and target states into a space of dimension $d_{share}$. This is also similar to the idea in \cite{zheng2015closed,zheng2017machine}, where different views of images use a shared regression coefficient. $(\alpha^T s_i^x - \alpha^T s_j^x)^2 W_x^{i,j}$ minimizes the difference of $s_i^x$ and $s_j^x$ with weight $W_x^{i,j}$. $W_x^{i,j}$ can be computed using the kernel function in Euclidean space:
\begin{align}
W_x^{i,j} = \exp(-\| s_i^x - s_j^x \|).  \label{eq:wx}
\end{align}
Similarly, $(\beta^T s_i^y - \beta^T s_j^y)^2 W_y^{i,j}$ minimizes the difference of $s_i^y$ and $s_j^y$ with weight $W_y^{i,j}$, which can be computed using Eq.(\ref{eq:wx}) similarly.

\begin{algorithm}[t]
\caption{Find projection $\chi$ using manifold alignment.}
\label{alg:alg2}
\begin{algorithmic}[1]
\Require Random source states $s^x_0, s^x_1,...$, random target states $s^y_0, s^y_1,...$
\Ensure Projection $\chi$ for state transfer $\mathcal{S}^x \to \mathcal{S}^y$.
\State Compute $W_x$,$W_y$,$W$ using Eqs.(\ref{eq:wx},\ref{eq:wij})
\State Compute $L_x$, $L_y$ using Eq.(\ref{eq:lx}).
\State Compute $\Omega_1$, $\Omega_2$, $\Omega_3$, $\Omega_4$ using Eq.(\ref{eq:om1}).
\State Formulate matrix $L$ and $Z$ using Eq.(\ref{eq:l}).
\State Solve Eq.(\ref{eq:ZLZ}).
\State Compute $\chi$ using Eq.(\ref{eq:chi_alg2}).
\end{algorithmic}
\end{algorithm}

$(\alpha^T s_i^x - \beta^T s_j^y)^2 W^{i,j}$ minimizes the difference of $s_i^x$ and $s_j^y$ in the shared space with weight $W^{i,j}$. To compute $W^{i,j}$, we can compare their knn local geometry matrix. Knn local geometry matrix $R_{s_i^x}$ is defined as a $(k+1) \times (k+1)$ matrix, with the $(k_1,k_2)$-th element as:
\begin{align}
R_{s_i^x}(k_1,k_2) = \| z_{k_1} - z_{k_2} \|, \label{eq:knn}
\end{align}
where $z = \{s_i^x, z_1, z_2, ..., z_k\}$, $z_{k_1}$ and $z_{k_2}$ are the $k_1$-th and $k_2$-th nearest neighbors, $\| z_{k_1} - z_{k_2} \|$ is Euclidean distance of $z_{k_1}$ and $z_{k_2}$. $R_{s_j^y}$ can be computed similarly using Eq.(\ref{eq:knn}). $W^{i,j}$ can then be given as:
\begin{align}
W^{i,j} &= \exp(-dist(R_{s_i^x}, R_{s_j^y}) ), \label{eq:wij}\\
dist(R_{s_i^x}, R_{s_j^y}) &= \min_{1\le h \le k!} min( dist_1(h), dist_2(h)), \label{eq:rij}
\end{align}
where $dist(R_{s_i^x}, R_{s_j^y})$ is the minimum distance of $k!$ possible permutations of the $k$ neighbors of $s_i^x$ and $s_i^y$. 
The distances of $h$-th permutation $dist_1(h)$ and $dist_2(h)$ is given as:
\begin{align}
dist_1(h) &= \| \{ R_{s_j^y} \}_h - w_1 R_{s_i^x} \|_F,\\
dist_2(h) &= \|R_{s_i^x}  - w_2  \{ R_{s_j^y} \}_h  \|_F,\\
w_1 &=  \Tr R_{s_i^x}^T  \{ R_{s_j^y} \}_h /\Tr R_{s_i^x}^T R_{s_i^x},\\
w_2 &=  \Tr  \{ R_{s_j^y} \}_h^T  R_{s_i^x} /\Tr \{ R_{s_j^y} \}_h^T \{ R_{s_j^y} \}_h,
\end{align}
where $\Tr$ is matrix trace operator, $w_1$ and $w_2$ are two weights terms. 

Minimizing cost function Eq.(\ref{eq:cost}) can be formulated as: 
\begin{align}
C(\phi) = \phi^T ZLZ^T \phi, \label{eq:ZLZ}
\end{align}
where 
\begin{align}
\phi = \begin{pmatrix} \alpha \\ \beta \end{pmatrix},
Z = \begin{pmatrix} X & 0 \\ 0 & Y \end{pmatrix}, 
L = \begin{pmatrix} L_x+\mu \Omega_1 & -\mu \Omega_2 \\ -\mu \Omega_3 & L_y+\mu \Omega_4 \end{pmatrix}. \label{eq:l}
\end{align}
Columns of $X$ are vector representations of state of $s^x_0, s^x_1,...$; columns of $Y$ are vector representations of state of $s^y_0, s^y_1,...$; $L_x$ and $L_y$ are given as:
\begin{align}
L_x = D_x - W_x, \; L_y = D_y - W_y,\label{eq:lx}
\end{align}
where $D_x$ and $D_y$ are diagonal matrix:
\begin{align}
D_x^{ii} = \sum_j W_x^{ij}, \; D_y^{ii} = \sum_j W_y^{ij}. 
\end{align}
$\Omega_1$ and $\Omega_4$ are diagonal matrices. $\Omega_2$ is the same as $W$, $\Omega_3$ is the same as the transpose of $W$. $\Omega_1$, $\Omega_2$, $\Omega_3$ and $\Omega_4$ are given as:
\begin{align}
\Omega_1^{ii} = \sum_j W^{ij},\; \Omega_2^{ij} =  W^{ij}, \; \Omega_3^{ij} = W^{ji}, \;\Omega_4^{ii} = \sum_j W^{ji}.\label{eq:om1}
\end{align}
$d_{share}$ is a tuning parameter indicating the size of the shared space. Finally, $\chi$ is given as:
\begin{align}
\chi = \beta^{T\dagger} \alpha^T, \label{eq:chi_alg2}
\end{align}
where $\dagger$ is matrix pseudo inverse.

\begin{algorithm}[t]
\caption{Recover policy $\pi^{y}$.}
\label{alg:alg3}
\begin{algorithmic}[1]
\Require State trajectories $\{ s^y_{00}, s^y_{01}, ... \}$, $\{ s^y_{10}, s^y_{11}, ... \}$,...
\Ensure Policy $\pi^{y}$.
\State Compute action trajectories in $\mathcal{A}^y$, $\{ a^y_{00}, a^y_{01}, ... \}$, $\{ a^y_{10}, a^y_{11}, ... \}$, ..., using Eq.(\ref{eq:ay}). 
\State Train $\pi^{y}$ with a deep model by using state trajectories as input feature, action trajectories as output class labels.
\end{algorithmic}
\end{algorithm}

\textbf{Recover policy $\pi^{y}$}. Algorithm \ref{alg:alg3} shows how to recover policy $\pi^{y}$ given state trajectory $\{ s^y_{00}, s^y_{01}, ... \}$, $\{ s^y_{10}, s^y_{11}, ... \}$,... We know that when at state $s^y_{0(t-1)}$, an action $a^y_{0(t-1)} \in \mathcal{A}^y$ was taken and then the state evolved from $s^y_{0(t-1)}$ to $s^y_{0t}$. However, we cannot directly get the action $a^y_{0(t-1)}$ by comparing $s^y_{0(t-1)}$ and $s^y_{0t}$. Our approach is to try all possible actions in $\mathcal{A}^y$ and record the states $\widetilde{s^y_{0t}}(a)$ after taking each action, then compare states $\widetilde{s^y_{0t}}(a)$ with $s^y_{0t}$. To compute $\widetilde{s^y_{0t}}(a)$, when action $a$ is taken, we remove the $a$-th job in job queue and set it to be $0$s in matrix $s^y_{0(t-1)}$. We use the following equation to find the $a^y_{0(t-1)}$:    
\begin{align}
a^y_{0(t-1)}= \argmin_{a\in \mathcal{A}^y} \| \widetilde{s^y_{0t}}(a) - s^y_{0t} \|, \label{eq:ay}
\end{align}
where $\| \widetilde{s^y_{0t}}(a) - s^y_{0t} \|$ is distance between states $\widetilde{s^y_{0t}}(a)$ and $s^y_{0t}$. 

Using Eq.(\ref{eq:ay}), we can find action trajectories $\{ a^y_{00}, a^y_{01}, ... \}$, $\{ a^y_{10}, $ $a^y_{11}, ... \}$, ... We initialize the target policy $\pi^{y}$ using a deep model, such as DNN or Deep CNN, and train policy $\pi^{y}$ as a classifier from $s^y$ to a corresponding $a^y$. 

\section{Experiments}
% In this section, we explore the performance of deep dispatching with different job characteristics and the efficiency of the proposed policy transfer algorithms. 

\subsection{Experiment setup and evaluation metrics}
\label{sectup}
\textbf{Experiment settings.}
Unless otherwise noted, the default setting is given as below: the machine looks $T=15$ time steps ahead, length of slack time array is $Z=5$, number of job slots is $n=10$, number of backlog slots is $m=60$. We test a set of different values for the following job characteristics parameters: 
\begin{enumerate}
  \item \textbf{Job arrival speed} $\lambda \in (0,1)$ is the probability of arriving a new job at each time step, with default $\lambda = 0.5$. 
  \item \textbf{Small job probability} $p_{small}$ indicates the probability that the new job length is short, with default $p_{small}=0.8$. The short job length distribution is a random number in $[1,2]$; the long job length distribution is a random number in $[6,10]$. 
  \item \textbf{Urgent job probability} $p_{urgent}$ indicates probability of the job slack time is very short, with default $p_{urgent}=0.5$. When the job arrives, the urgent job slack time distribution is a random number in $[1,5]$; the non-urgent job slack time distribution is a random number in $[5,10]$. 
\end{enumerate}

At any time step, if the job queue is full, the new job will be dropped, a penalty $-10$ will be added to reward $r_t$. We use Deep Neural Network (DNN) to model policy $\pi_{a|s;\theta}$. The input is state representation from Figure \ref{fig:image}. The output is a probability vector of length same as action space $\mathcal{A}$. The activation function in hidden layer is rectified linear unit and the activation function in the last layer is Softmax. The DNN network structure we use is 2 hidden layers with 128 and 64 neurons respectively. 

\textbf{Training and testing data.}
For each experiment, we use the same setting to generate 10 random trajectories for training and testing respectively. Then we train each comparing method on the same training data and test it on the same testing data. 

\textbf{Evaluation metrics.}
To evaluate the performance and business impact, we use total discounted reward (Eq.(\ref{eq:j})) and average lateness (Eq.(\ref{eq:lateness})), tardiness (Eq.(\ref{eq:tardiness})). Total discounted reward includes the penalty when a job was dropped if the job queue is full. Average lateness and tardiness only consider those jobs that are successfully finished, which are straightforward and valuable to business users. The reported results are the average values over 10 random testing trajectories.

\begin{figure}[t]
\centering
\begin{subfigure}[b]{0.3\columnwidth}
        \includegraphics[width=\textwidth]{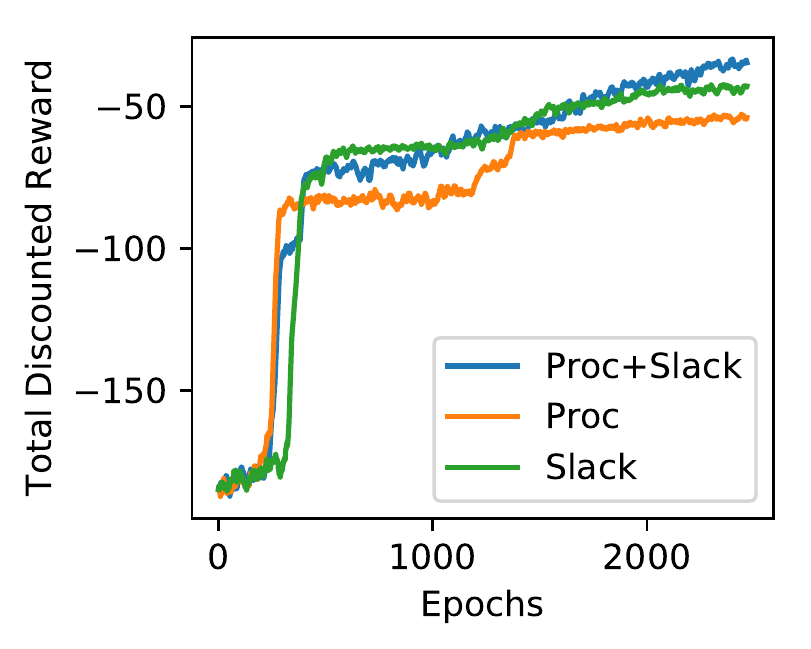}
        \caption{Total reward.}
        \label{fig:}
\end{subfigure}
\begin{subfigure}[b]{0.28\columnwidth}
        \includegraphics[width=\textwidth]{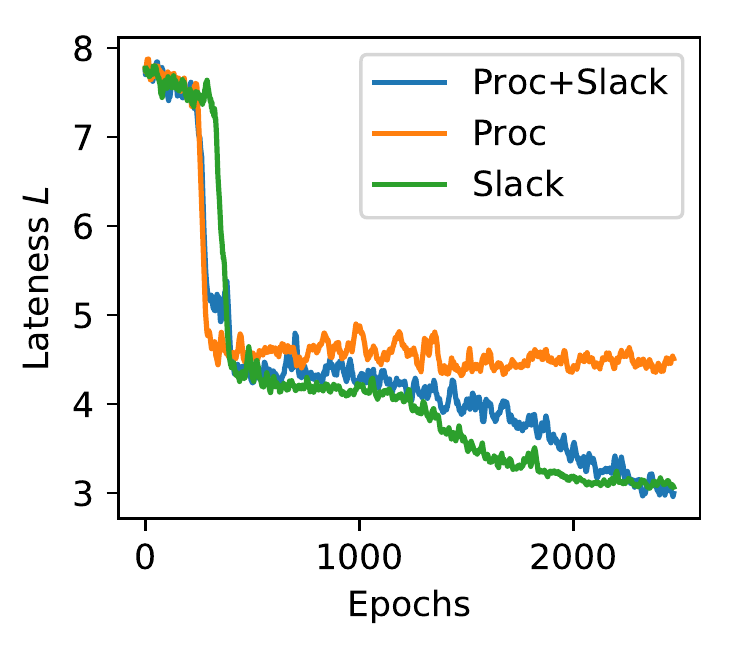}
        \caption{Average lateness.}
        \label{fig:}
\end{subfigure}
\caption{Effectiveness of state presentation. ``Proc+Slack": using both Processing time and Slack time of Job queue in Figure \ref{fig:image}. ``Proc": using only Processing time. ``Proc": using only Slack time.}
\label{fig:slack}
\end{figure}

\subsection{Effectiveness of state presentation Figure \ref{fig:image}}
Figure \ref{fig:slack} shows the effectiveness of combining processing time and slack time information in state presentation (Figure \ref{fig:image}). Combining both processing and slack time gives the highest total discounted lateness reward and lowest average lateness. Different to computer job resource management \cite{mao2016resource}, slack time is important for manufacturing dispatching.

\begin{table}[h!]
\centering
\caption{Total discounted reward using different objective and trajectory length.}
\label{tab:tab1}
\resizebox{0.8\columnwidth}{!} {
\begin{tabular}{l|rrrr|rrrr}
\hline
\hline
                 & \multicolumn{4}{c|}{Lateness objective}               & \multicolumn{4}{c}{Tardiness objective}  \\
 \hline
Trajectory length   & 50 & 100 & 150 & 200 & 50&100 & 150 & 200 \\
\hline
EDF              & -65.05                 & -116.06                 & -215.65                 & -513.53                 & -57.03                 & -95.67                  & -192.64                 & -493.14                 \\
LST              & -77.11                 & -144.45                 & -260.93                 & -624.76                 & -69.09                 & -130.97                 & -247.45                 & -611.28                 \\
Random Forest	&-65.05&	-115.47&	-211.37&	-433.40&		-57.03&	-92.46&	-195.20&	-410.39\\
SVM	            &-65.05&	-115.47&	-212.48&	-348.01&		-57.03&	-92.46&	-185.50&	-324.99\\
Neural Network	&-65.05&	-115.47&	-164.58&	-238.31&		-57.03&	-92.46&	-137.60&	-199.19 \\
Tetris           & -21.65                 & -89.37                  & -73.66                  & -154.15                 & -8.93                  & -16.67                  & -24.46         & -102.53        \\
RL-Mao           & -7.00                  & -57.39                  & -250.51                 & -212.32                 & -38.86                 & -118.54                 & -223.05                 & -370.97                 \\
DMD & \textbf{-0.51}         & \textbf{ -34.47}         & \textbf{-63.30}         & \textbf{-133.14}        & \textbf{-0.22}         & \textbf{-4.09}          & \textbf{-9.43}                  & \textbf{-14.55}                \\
\hline
\hline
\end{tabular}
}
\centering
\caption{Average lateness and tardiness using different objective and trajectory length.}
\label{tab:tab2}
\resizebox{0.6\columnwidth}{!} {
\begin{tabular}{l|rrrr|rrrr}
\hline
\hline
                 & \multicolumn{4}{c|}{Average lateness}               & \multicolumn{4}{c}{Average tardiness}  \\
 \hline
Trajectory length    & 50 & 100 & 150 & 200 & 50&100 & 150 & 200 \\
\hline
EDF              & 5.29          & 4.44          & 7.02          & 11.68         & 5.77          & 5.38          & 8.71          & 13.89         \\
LST              & 6.71          & 5.48          & 8.26          & 13.80         & 7.07          & 6.21          & 9.31          & 15.35         \\
Random Forest&	5.29&	4.38&	5.89&	9.03&		5.77&	5.40&	6.69&	10.66\\
SVM&	        5.29&	4.38&	5.48&	7.32&		5.77&	5.40&	6.72&	8.70\\
Neural Network&	5.29&	4.38&	4.37&	5.23&		5.77&	5.40&	5.29&	7.33\\
Tetris           & 6.13          & 8.08          & 5.73          & 8.20          & 8.25          & 14.17         & 6.20          & 10.07         \\
RL-Mao           & 5.29          & 6.39          & 7.35          & 8.06          & 7.78          & 9.10          & 9.21          & 10.43         \\
DMD & \textbf{2.11} & \textbf{3.16} & \textbf{3.60}          & \textbf{5.01} & \textbf{1.51} & \textbf{2.24} & \textbf{3.68}          & \textbf{4.14}         
\\
\hline
\hline
\end{tabular}
}
%\end{table}
%\begin{table*}[t]
\centering
\caption{Total discounted lateness reward with various job characteristics.}
\label{tab:tab3}
\resizebox{1\textwidth}{!} {
\begin{tabular}{l|rrrrr|rrr|rrr}
\hline
\hline
    & \multicolumn{5}{c|}{Job arrival speed $\lambda$}                                                    & \multicolumn{3}{c|}{Small job $p_{small}$}   & \multicolumn{3}{c}{Urgent job $p_{urgent}$} \\
\hline
                 & 0.1           & 0.3           & 0.5           & 0.7           & 0.9           & 0.2           & 0.5           & 0.8           & 0.1           & 0.5           & 0.9           \\
\hline
EDF              & -31.12          & -96.99          & -116.06         & -317.59         & -415.87          & -305.49         & -246.97         & -116.06         & -71.92          & -116.06         & -115.08         \\
LST              & -31.12          & -126.23         & -144.45         & -378.46         & -422.32          & -316.67         & -278.47         & -144.45         & -75.77          & -144.45         & -120.12         \\
Random Forest&	-31.12&	-78.22&	-115.47&	-205.24&	-246.33&		-106.16&	-106.33&	-115.47&		-71.15&	-115.47&	-111.95\\
SVM&	        -31.12&	-78.66&	-115.47&	-208.10&	-131.94&		-154.18&	-124.25&	-115.47&		-71.15&	-115.47&	-111.95\\
Neural Network&	-31.12&	-78.66&	-115.47&	-219.09&	-260.66&		-111.56&	-150.48&	-115.47&		-71.15&	-115.47&	-111.95 \\
Tetris           & -28.39          & -83.46          & -89.37          & \textbf{-53.08} & -107.22 & -70.08          & -48.26          & -89.37          & -68.74          & -89.37          & -69.85          \\
RL-Mao           & -13.79          & -58.95          & -57.39          & -136.36         & -178.74          & -99.27          & -74.24          & -57.39          & -49.70          & -57.39          & -48.11          \\
DMD & \textbf{-11.12} & \textbf{-14.48} & \textbf{ -34.47} & -62.37          & \textbf{-106.70}          & \textbf{-56.33} & \textbf{-43.26} & \textbf{ -34.47} & \textbf{-38.45} & \textbf{ -34.47} & \textbf{-36.86}\\
\hline
\hline
\end{tabular}
}
%\end{table*}
%\begin{table}[t]
\centering
\caption{Average lateness with various job characteristics.}
\label{tab:tab4}
\resizebox{0.7\textwidth}{!} {
\begin{tabular}{l|rrrrr|rrr|rrr}
\hline
\hline
    & \multicolumn{5}{c|}{Job arrival speed $\lambda$}                                                    & \multicolumn{3}{c|}{Small job $p_{small}$}   & \multicolumn{3}{c}{Urgent job $p_{urgent}$} \\
\hline
                 & 0.1           & 0.3           & 0.5           & 0.7           & 0.9           & 0.2           & 0.5           & 0.8           & 0.1           & 0.5           & 0.9           \\
\hline
EDF              & 5.33          & 5.00          & 4.44          & 16.29         & 19.41         & 21.05         & 12.83         & 4.44          & 3.26          & 4.44          & 4.67          \\
LST              & 5.33          & 6.28          & 5.48          & 20.89         & 22.54         & 21.18         & 15.00         & 5.48          & 4.00          & 5.48          & 4.94          \\
Random Forest&	5.33&	4.00&	4.38&	8.21&	12.73&		8.11&	5.42&	4.38&		\textbf{2.88}&	4.38&	4.32\\
SVM	&           5.33&	4.16&	4.38&	8.21&	\textbf{4.02}&	9.47&	5.68&	4.38&		\textbf{2.88}&	4.38&	4.32\\
Neural Network&	5.33&	4.16&	4.38&	10.30&	15.43&		\textbf{7.61}&	6.36&	4.38&		\textbf{2.88}&	4.38&	4.32\\
Tetris           & 5.33          & 7.67          & 8.08          & 6.58          & 12.78         & 7.75          & 6.30          & 8.08          & 7.68          & 8.08          & 6.26          \\
RL-Mao           & 4.50          & 6.65          & 6.39          & 9.35          & 10.87         & 11.62         & 8.19          & 6.39          & 5.15          & 6.39          & 5.83          \\
DMD & \textbf{3.45} & \textbf{3.21} & \textbf{3.16} & \textbf{4.51}  & 9.74         & 7.70 & \textbf{4.32} & \textbf{3.16} & 4.55          & \textbf{3.16} & \textbf{4.03} \\
\hline
\hline
\end{tabular}
}
\end{table}

\subsection{Comparison with existing algorithms}
We compare the proposed Deep Manufacturing Dispatching (DMD) with 7 other dispatching policies: 2 due time related manual designed rules (EDF for Earliest-Due-First, LST for Least-Slack-Time), 3 hyper-heuristics using machine learning (random forest, SVM, and neural network with 2 hidden layers), and reinforcement learning based RL-Mao \cite{mao2016resource} and Tetris \cite{grandl2015multi}. For hyper-heuristics, under each experiment setting, we choose the best heuristic rule in terms of average lateness or tardiness as ground truth rule. 

Table \ref{tab:tab1} and \ref{tab:tab2} show DMD gets the highest reward and lowest average lateness and tardiness. Unless other noted, the default trajectory length is $100$ time steps. In manufacturing scheduling, where job frequency is not as high as computer jobs, this trajectory length is reasonable. We can see from Table \ref{tab:tab1} that as the length of trajectory increases, the total discounted reward decreases overall. Table \ref{tab:tab3} and \ref{tab:tab4} show total discounted reward and average lateness with different settings of $\lambda$, $p_{small}$ and $p_{urgent}$. Overall, for 19 settings (8 settings in Tables \ref{tab:tab1},\ref{tab:tab2} and 11 settings in Tables \ref{tab:tab3},\ref{tab:tab4}), DMD gets best results for 18 settings on total discounted reward and 16 settings on average lateness and tardiness. Best performing methods in each column are in bold.

\begin{figure}[t]
\centering
\begin{subfigure}[b]{0.25\columnwidth}
        \includegraphics[width=\textwidth]{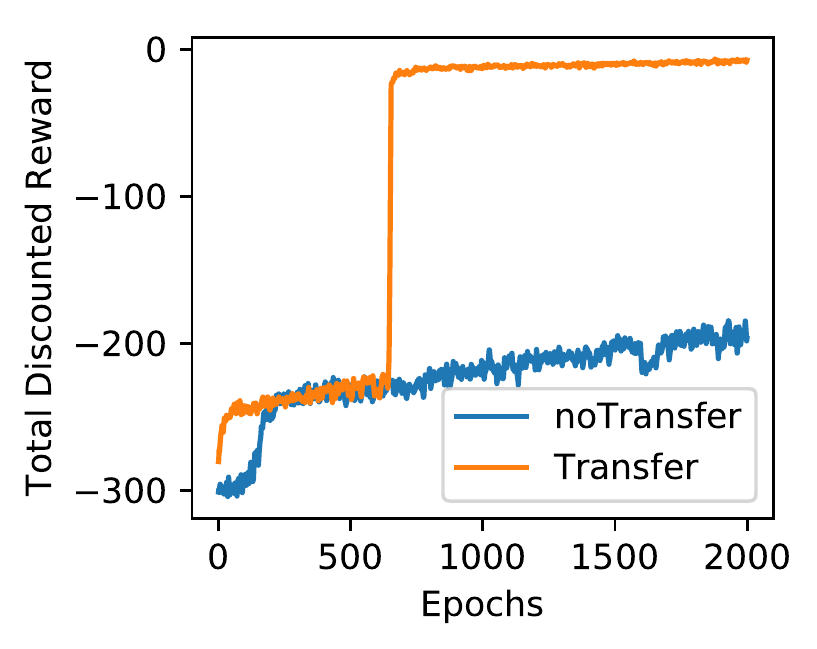}
        \caption{Setting 1.}
        \label{fig:}
\end{subfigure}%
\begin{subfigure}[b]{0.25\columnwidth}
        \includegraphics[width=\textwidth]{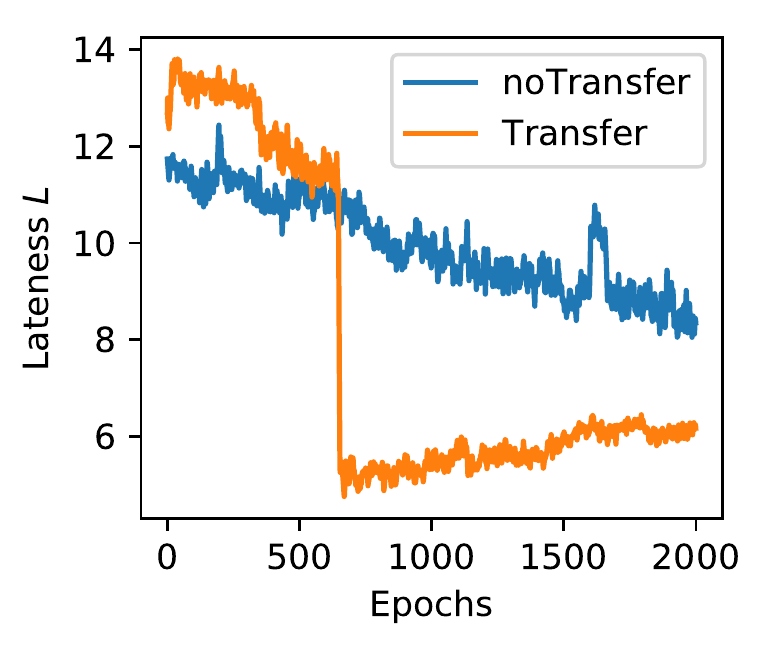}
        \caption{Setting 1.}
        \label{fig:}
\end{subfigure}%
\begin{subfigure}[b]{0.25\columnwidth}
        \includegraphics[width=\textwidth]{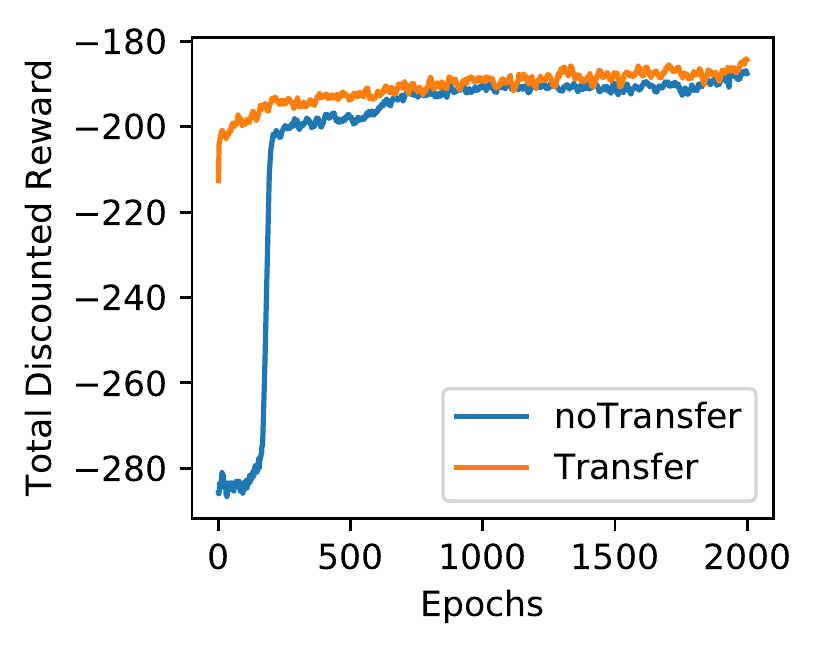}
        \caption{Setting 2.}
        \label{fig:}
\end{subfigure}%
\begin{subfigure}[b]{0.25\columnwidth}
        \includegraphics[width=\textwidth]{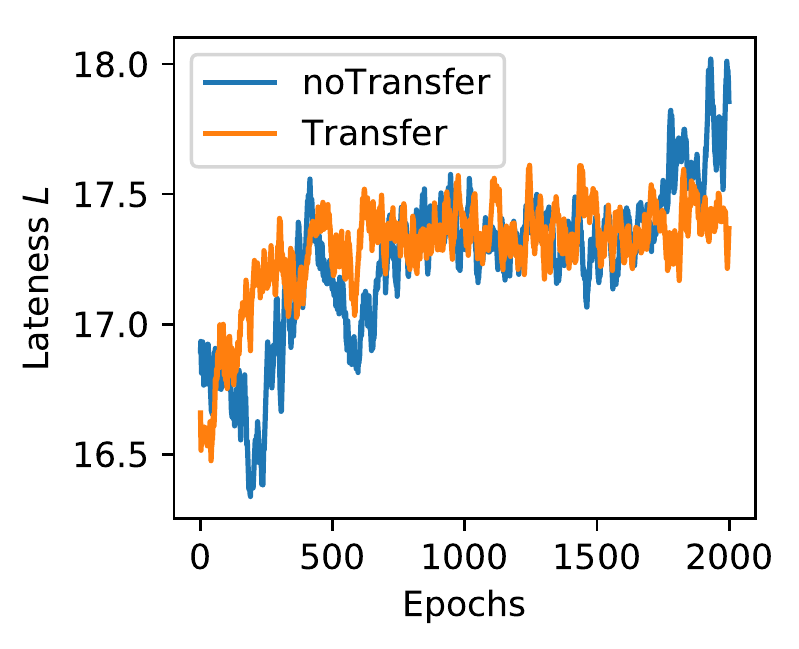}
        \caption{Setting 2.}
        \label{fig:}
\end{subfigure}\\
\begin{subfigure}[b]{0.25\columnwidth}
        \includegraphics[width=\textwidth]{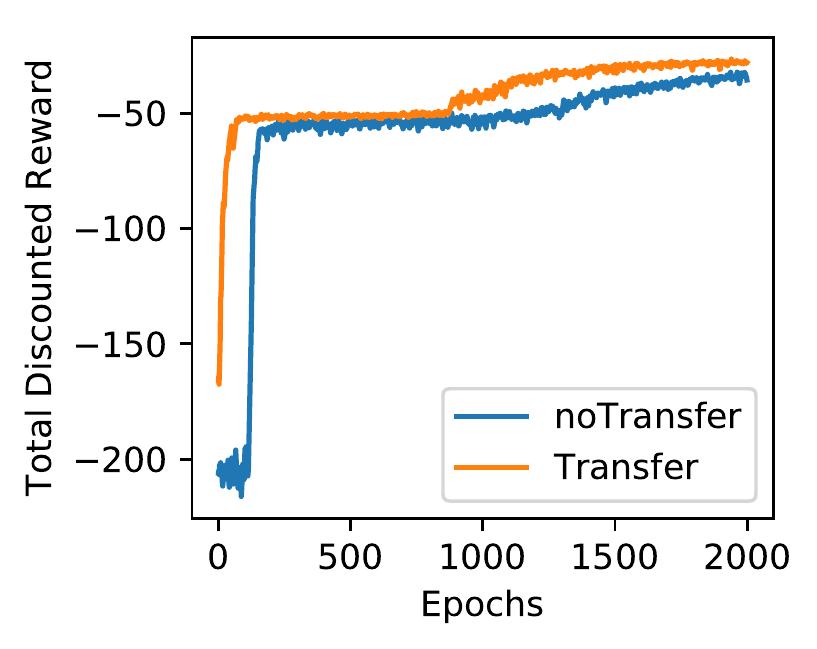}
        \caption{Setting 3.}
        \label{fig:}
\end{subfigure}%
\begin{subfigure}[b]{0.25\columnwidth}
        \includegraphics[width=\textwidth]{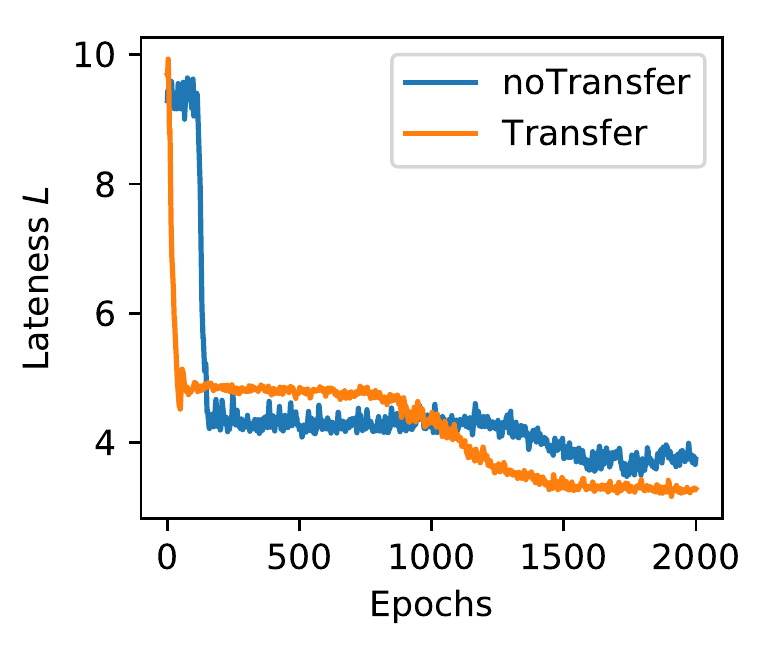}
        \caption{Setting 3.}
        \label{fig:}
\end{subfigure}%
\begin{subfigure}[b]{0.25\columnwidth}
        \includegraphics[width=\textwidth]{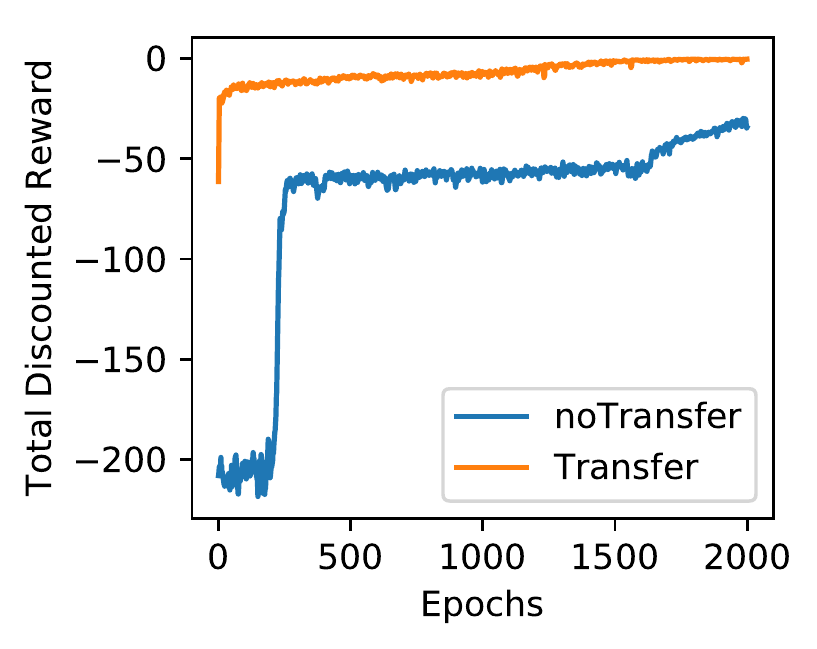}
        \caption{Setting 4.}
        \label{fig:}
\end{subfigure}%
\begin{subfigure}[b]{0.25\columnwidth}
        \includegraphics[width=\textwidth]{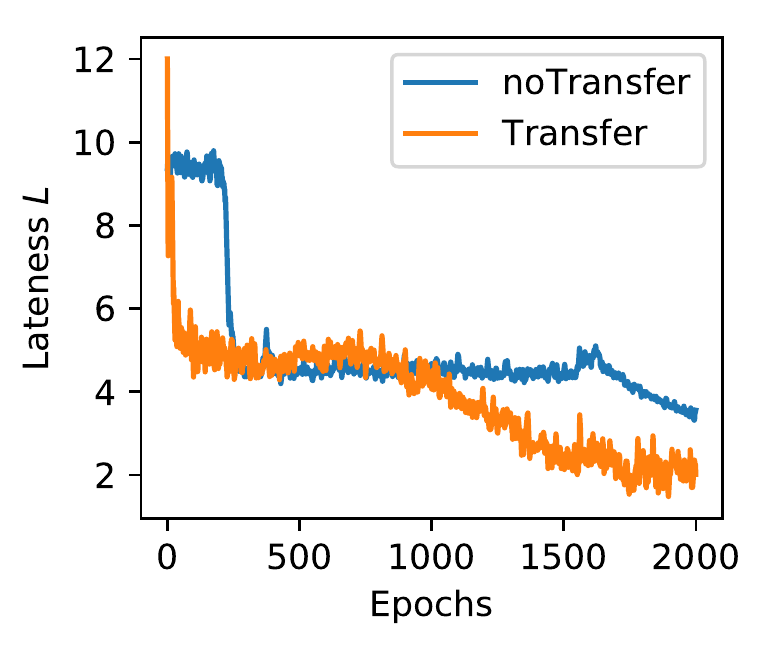}
        \caption{Setting 4.}
        \label{fig:}
\end{subfigure}
\caption{Policy transfer evaluation using Deep Manufacturing Dispatching (DMD).}
\label{fig:transfer}
\end{figure}

\begin{table}[t]
\centering
\caption{Policy transfer evaluation for hyper-heuristics.}
\label{tab:tab5}
\resizebox{0.9\textwidth}{!} {
\begin{tabular}{l|l|rrrr|rrrr}
\hline
\hline
  &   & \multicolumn{4}{c|}{noTransfer}  & \multicolumn{4}{c}{Transfer}  \\
  \hline
Transfer setting &   & 1 & 2 & 3 &4  & 1 & 2 & 3 &4  \\
\hline
\multirow{3}{*}{\begin{tabular}[c]{@{}l@{}}Total\\ discounted \\ lateness\end{tabular}} & Random Forest  & -205.24 & -216.44          & -120.22          & -120.22          & \textbf{-140.02} & \textbf{-205.97} & \textbf{-115.47} & \textbf{-115.47} \\
                                                                                        & SVM            & -208.10 & \textbf{-189.27} & \textbf{-115.47} & \textbf{-115.47} & \textbf{-128.26} & -213.40          & \textbf{-115.47} & \textbf{-115.47} \\
                                                                                        & Neural Network & -219.09 & -229.46          & \textbf{-114.15} & -121.65          & \textbf{-174.52} & \textbf{-204.55} & -115.47          & \textbf{-115.47} \\
\hline
\multirow{3}{*}{\begin{tabular}[c]{@{}l@{}}Average\\ lateness\end{tabular}}             & Random Forest  & 8.21    & 14.05            & 4.63             & 4.63             & \textbf{5.08}    & \textbf{10.63}   & \textbf{4.38}    & \textbf{4.38}    \\
                                                                                        & SVM            & 8.21    & 13.05            & \textbf{4.38}    & \textbf{4.38}    & \textbf{5.13}    & \textbf{10.96}   & \textbf{4.38}    & \textbf{4.38}    \\
                                                                                        & Neural Network & 10.30   & 16.74            & \textbf{4.38}    & 4.72             & \textbf{7.00}    & \textbf{13.40}   & \textbf{4.38}    & \textbf{4.38}\\
\hline
\hline
\end{tabular}
}
\end{table}

\subsection{Dispatching policy transfer}
In step 1 of Algorithm \ref{alg:alg1}, we generate 2000 random states from source environment and target environment respectively for each transfer setting. The following four transfer learning settings are considered:
(1). Same-environment: Source $\lambda=0.5$; Target $\lambda=0.7$.
(2). Same-environment: Source $\lambda=0.5$, $p_{small}=0.8$; Target $\lambda=0.7$, $p_{small}=0.5$.
(3). Cross-environment: Source $n=10$; Target $n=15$.
(4). Cross-environment: Source $n=10$, $m=60$; Target $n=15$, $m=30$.
In all figures, \emph{noTransfer} curve is trained from scratch under target setting, \emph{Transfer} curve is initialized using the output of Algorithm \ref{alg:alg3} under source setting, then trained under target setting. Figure \ref{fig:transfer} shows total discounted reward and average lateness of the 4 transfer settings using DMD. We find that policy transferring reduces training time and gives larger total discounted reward and smaller average lateness compared to training from scratch. Table \ref{tab:tab5} shows the results using 3 hyper-heuristics. Best results are highlighted for each transfer setting. Note that, as mentioned in Section \ref{sectup}, dropped job penalty is considered in total discounted reward, but not in average lateness. We find that policy transfer gives competitive or better results than training from scratch using hyper-heuristics. This shows the effectiveness of Algorithm \ref{alg:alg1}.

\section{Conclusion}
Dispatching is a difficult yet important problem. It is not straightforward due to the complexity of manufacturing settings. We showed the promising application of combining reinforcement learning and transfer learning in manufacturing industry. The transfer learning module increases the generalization of learned dispatching rules and saves cost and time for data collection and model training.

%\bibliographystyle{splncs04}
%\bibliography{bib1}

\end{document}